\documentclass[letterpaper]{IEEEtran}

\usepackage{cite}
\usepackage{amsmath,amssymb,amsfonts}
\usepackage{gensymb}
\usepackage{textcomp}
\usepackage{xcolor}

\usepackage{algorithm,algpseudocode}

\usepackage{flushend}
\usepackage{listings}
\DeclareMathOperator*{\argmax}{argmax} % thin space, limits underneath in displays
\usepackage{mathptmx}
\usepackage{graphicx}
\usepackage{textcomp}
\def\BibTeX{{\rm B\kern-.05em{\sc i\kern-.025em b}\kern-.08em
    T\kern-.1667em\lower.7ex\hbox{E}\kern-.125emX}}
\usepackage{makecell}

\usepackage[english]{babel}
\usepackage{amsthm}
\theoremstyle{definition}

\usepackage{caption}
\usepackage{subcaption}
\usepackage{mwe}

\begin{document} 
% Suboptimal Multi-Sensors Deployment for Barrier Coverage in Detecting Poisson-Distributed Target Trajectories

\title{Near-optimal Sensor Placement for Detecting Stochastic Target Trajectories in Barrier Coverage Systems
% Sensor Location Selection when Detecting Poisson-Distributed Target Trajectories with a Barrier Coverage System
\thanks{This work was supported in part by the Office of Naval Research under Grant N00014-20-1-2845.}
}

\author{Mingyu Kim$^{1*}$, Daniel J. Stilwell$^1$, Harun Yetkin$^1$, and Jorge Jimenez$^2$ \\ $^1$Electrical and Computer Engineering,
Virginia Polytechnic Institute and State University, Blacksburg, VA, USA\\
% $^2$ Mechatronics Engineering,  Bart{\i}n University, Turkey \\
$^2$ Johns Hopkins University Applied Physics Laboratory, Laurel, MD, USA\\
$^*$mkim486@vt.edu}

% make the title area
\maketitle

\begin{abstract}
% This paper addresses a 2-D barrier coverage problem  where we search for suboptimal sensor locations for detection of targets traveling  with their trajectories modeled by a log-Gaussian Cox line process. We investigate sensor placement in the corresponding representation space where linear target trajectories are represented points. While the line process is more easily addressed in the representation space, the spatial functions that represent sensor performance (e.g., probability of detection) are less intuitive in the representation space. To demonstrate our approach, we consider the problem of choosing the location of sensors that are placed and the seafloor and detect passing ships. We conduct numerical experiments utilizing historical ship data and compute sensor locations that minimize the number of undetected ships.

This paper addresses the deployment of sensors for a 2-D barrier coverage system. The challenge is to compute near-optimal sensor placements for detecting targets whose trajectories follow a log-Gaussian Cox line process. We explore sensor deployment in a transformed space, where linear target trajectories are represented as points. While this space simplifies handling the line process, the spatial functions representing sensor performance (i.e. probability of detection) become less intuitive. To illustrate our approach, we focus on positioning sensors of the barrier coverage system on the seafloor to detect passing ships. Through numerical experiments using historical ship data, we compute sensor locations that maximize the probability all ship passing over the barrier coverage system are detected.

\end{abstract}


\IEEEpeerreviewmaketitle

\section{Introduction}
\label{sec:introduction}

Barrier coverage systems have been widely studied in various multi-agent system applications, such as unmanned aerial vehicles (UAVs) and sensor networks. In these scenarios, devices are deployed to create a coverage area that detects targets within a specified region. These applications include wildlife monitoring \cite{hammoudeh2017wireless, tao2014survey, garcia2010wireless}, pest control \cite{si2022target}, search-and-rescue operations \cite{han2020scheduling}, and border surveillance \cite{benahmed2019optimal, aranzazu2019uav, mostafaei2018border, hammoudeh2017wireless}. The objectives of barrier coverage can vary depending on the application, often focusing on minimizing energy consumption or maximizing detection coverage.

Despite extensive research, most studies do not account for uncertain target and probabilistic sensor performance. Typically, many agents are deployed to ensure that an area of interest is fully protected \cite{hammoudeh2017wireless, han2020scheduling, si2022target}. While this approach can be effective, it may require a larger number of sensors than necessary in practice. In this paper, we aim to improve sensor deployment efficiency by incorporating a probabilistic target model based on historical target data. We seek near-optimal sensor locations for a fixed but arbitrary number of sensors. This allows for a more strategic deployment of a finite number of sensors, rather than attempting to cover the entire area with numerous sensors.

To model target behavior, we employ a stochastic process that captures the uncertainty of target arrivals, which is crucial for realistically modeling real-world intrusions. Informed by historical data, this approach provides a more accurate representation of target dynamics. While most barrier coverage studies disregard uncertainty in target modeling, there are some notable exceptions. The authors in \cite{aranzazu2019uav, he2013mobility} model target arrivals in 1D space using homogeneous point processes, such as Weibull or Poisson distributions with predefined intensities. However, for practical applications, especially when historical data is available, a nonhomogeneous Poisson process with an uncertain intensity function offers a more realistic model. In our previous work \cite{kim2023toward}, we utilize a log-Gaussian Cox process (LGCP) to model target arrivals for a 1D barrier coverage system, establishing a framework to identify suboptimal sensor locations in real-time to maximize void probability (i.e., the probability that all targets are detected).

In this paper, we extend the framework in \cite{kim2023toward} for a 2D barrier coverage system focused on detecting stochastic linear target trajectories rather than arrival locations for 1D barrier coverage system in \cite{kim2023toward}. This model is well-suited for small domains where target paths can be approximated as linear, though the stochastic nature of these trajectories introduces complexities in expressing their interaction with sensor performance in 2D space. Additionally, estimating these stochastic trajectories in the original domain poses challenges.

To overcome these, we apply a bijective transformation that maps linear target trajectories in 2D to distinct points in a new domain. Our objective in this barrier coverage problem is to minimize the probability of undetected target trajectories, effectively maximizing the void probability (i.e., probability of perfect detection). Building upon our previous work \cite{kim2023toward}, this framework optimizes sensor placement and performance within this transformed domain.

Our approach maps target trajectories, represented by a Poisson line process, to a representation space $\mathcal{C}$, treating the line process as a point process. Sensor performance, typically modeled as a spatial function of detection probability based on proximity to the sensor, is also mapped to this representation space. We select sensor locations by solving an optimization problem within the representation space. Specifically, linear trajectories in the original coordinate system, known as the inertial space, are mapped to unique points in the representation space \cite{chiu2013stochastic}. Target trajectories are modeled using a log-Gaussian Cox line process (LGCLP), and the intensity function of the line process is estimated in the representation space. Finally, a modified greedy algorithm from our prior work \cite{kim2023toward} is applied to efficiently locate near-optimal sensor placements in this transformed space.

Poisson processes, widely used to model spatial targets in various fields such as crime rate modeling \cite{park2021investigating, assunccao2007space, shirota2017space} and disease mapping \cite{diggle2005point, choiruddin2023covid, diggle2006spatio}, often assume static target locations. In contrast, ecological studies estimating animal populations (e.g., endangered species \cite{bowler2019integrating, wang2020genetic, schofield2020long}, marine mammals \cite{hodgson2017unmanned, fregosi2022detection}) typically use distance sampling methods like line-transect or point-transect sampling, assuming either that the targets are motionless or that their speed is negligible compared to the observer. These studies often rely on short-term observations, referred to as \emph{snapshots}. Hodgson et al. \cite{hodgson2017unmanned} and Fregosi et al. \cite{fregosi2022detection} show that neglecting target motion can significantly overestimate population sizes due to redundant detections. 

While no prior research addresses sensor location selection specifically for targets modeled as a Cox line process, Poisson line processes are widely applied in other contexts. For instance, the authors in \cite{chetlur2018coverage, choi2018poisson} model road networks as Poisson line processes, with nodes on these networks represented as Cox processes.

\subsection*{Contributions}

To the best of our knowledge, this work is the first to address 2D barrier coverage systems with sensors detecting targets modeled by a log-Gaussian Cox line process. We rigorously examine the interaction between continuous sensor performance and stochastic linear target trajectories within a representation space, enabling an efficient search for near-optimal sensor placements. Our framework optimizes sensor locations to maximize the probability of detecting all targets, offering practical insights into numerical optimization techniques for sensor deployment.

Our approach includes: (1) mapping historical target trajectory data into a representation space as a Cox point process realization, (2) estimating the Cox point process intensity function via the integrated nested Laplace approximation (INLA), (3) linking sensor performance to the representation space to select sensor locations that optimally thin the Cox process, (4) implementing an iterative optimization algorithm, initialized with a greedy sensor selection, and (5) transforming the final sensor placements back to the inertial space. We illustrate this approach using numerical examples based on historical ship traffic data from the Automated Identification System (AIS) near Hampton Roads Channel, Virginia, showing the model's practical application in real-world scenarios.

While the sensor performance function may be less intuitive in the representation space, it can be interpreted as thinning the target intensity function, which aids in optimal sensor placement. This framework can be applied to a wide range of sensor localization problems. For non-linear trajectories, alternative parameterizations may be used, though we assume the observation area is small enough for a linear approximation to suffice.

The paper is organized as follows: In Section \ref{sec: problem formulation}, we introduce the log-Gaussian Cox process for modeling target trajectories and define the corresponding sensor model in the representation space. Section \ref{sec:sensorplacement} outlines a computational approach to maximize target detection using the models from Section \ref{sec: problem formulation}. Section \ref{sec: numerical result} presents numerical results using historical ship traffic data.  Finally, Section \ref{sec: conclusion} concludes the paper.

% The appendix contains the proof of the bijective transformation used to map linear trajectories to unique points, as discussed in Section \ref{sec: problem formulation}.

%%%%%%%%%%%%%%%%%%%%%%%%%%%%%%%%%%%%%%%%%%%%%%%%%%%%%%%%%%%%%%%%%%%%%%%%%%%%%%%%%%%%%%%%%%%%%%%%%%%%%
\begin{figure}[b!] % figure 1
\centering
\includegraphics[scale=0.45]
{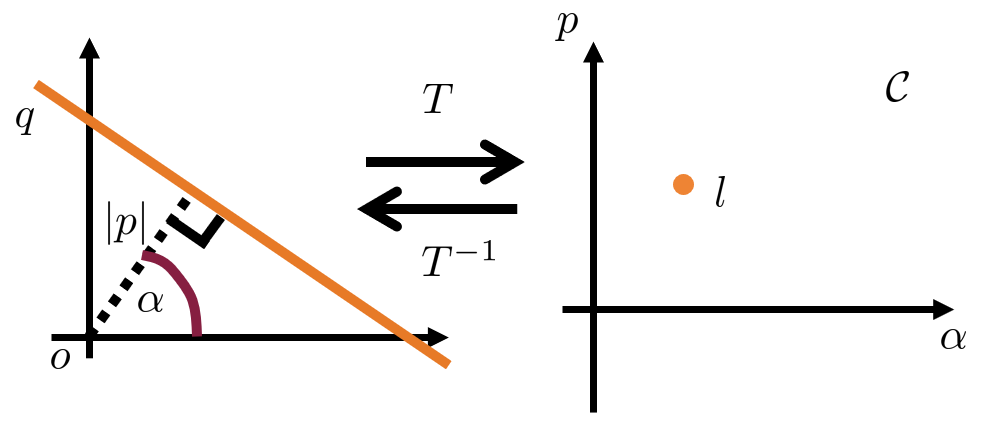}
\caption{(top) Linear trajectory $q$ is in the inertial space. (bottom) The corresponding mapped point $l$ in the representation space.
}\label{fig:1}
\end{figure}
%%%%%%%%%%%%%%%%%%%%%%%%%%%%%%%%%%%%%%%%%%%%%%%%%%%%%%%%%%%%%%%%%%%%%%%%%%%%%%%%%%%%%%%%%%%%%%%%%%%%%%
%%%%%%%%%%%%%%%%%%%%%%%%%%%%%%%%%%%%%%%%%%%%%%%%%%%%%%%%%%%%%%%
\begin{figure}[t!] % figure 2
\centering
\includegraphics[scale=0.3]
{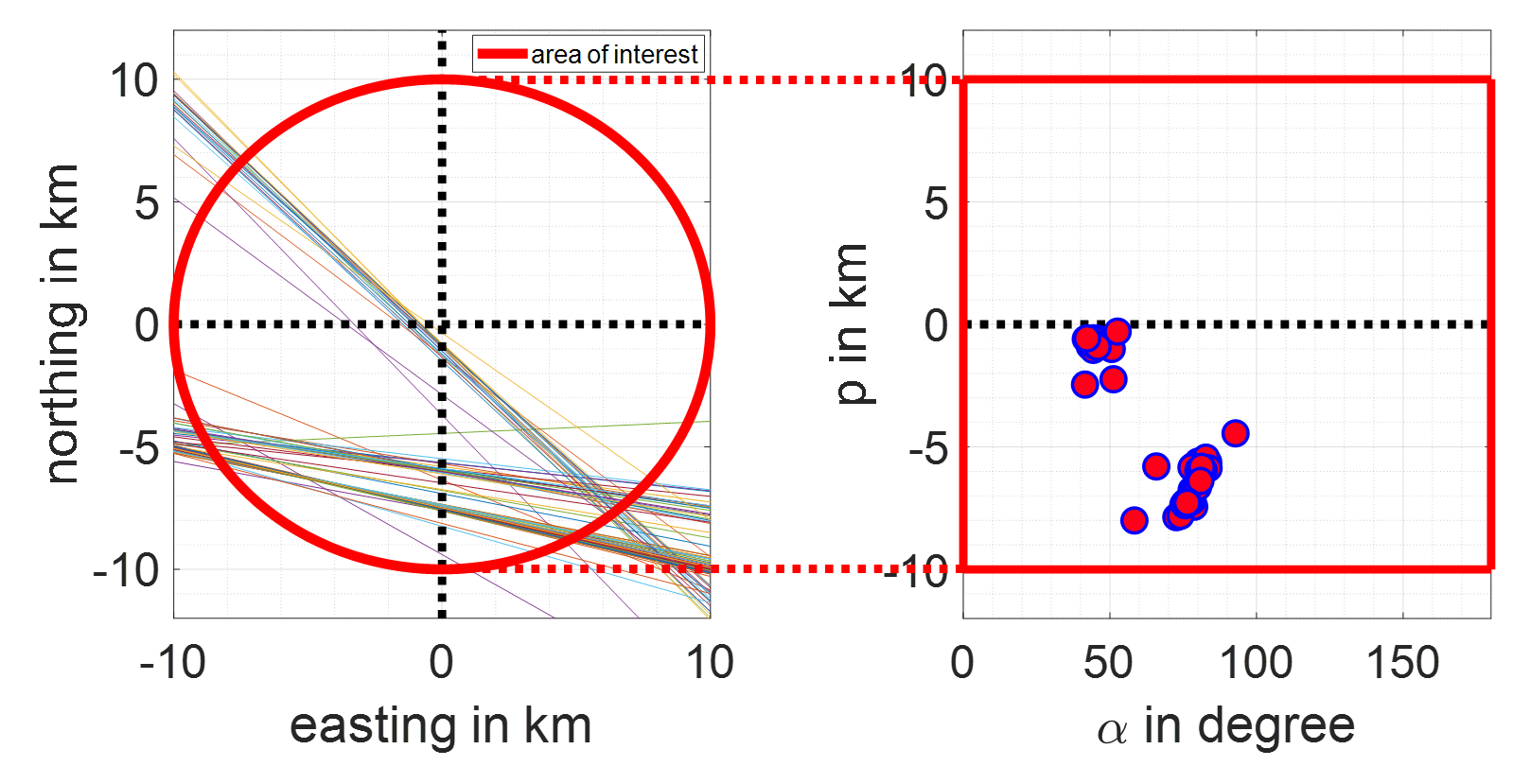}
\caption{(left) Estimated linear target trajectories passing through an area of interest. (right) The corresponding mapped points and the area of interest in the representation space.
}\label{fig:2}
\end{figure}
%%%%%%%%%%%%%%%%%%%%%%%%%%%%%%%%%%%%%%%%%%%%%%%%%%%%%%%%%%%%%%%%%%%%%%%%%%%%%%%%%%%%%%%%%%%%%%%%%%%%%%

\section{Problem Formulation \\ in a Representation Space $\mathcal{C}$ }
\label{sec: problem formulation}

We estimate the intensity function of the line process in the representation space $\mathcal{C} = { (\alpha, p): \alpha \in [0, \pi), p \in \mathbb{R} }$. This representation is standard, as seen in Section 8 of \cite{chiu2013stochastic}. For a given line $q$, $\alpha$ represents the angle between the line and the horizontal axis, while the magnitude of $p$ is the shortest distance between the line and the origin. If the line intercepts the negative vertical axis, $p$ is negative; otherwise, $p$ is non-negative. Additionally, if $q$ is parallel to the vertical axis, $p$ is positive if the trajectory is in the right-half plane, and non-positive if it is in the left-half plane. The value of $p$ can be computed 

\begin{align}
    p = 
\begin{cases}
    +\textit{$\left| p \right|$},& \text{if } \alpha = 0 \text{ and $q$ is in the closed RHP} \\
    -\textit{$\left| p \right|$},& \text{if } \alpha = 0 \text{ and $q$ is in the open LHP} \\
    +\textit{$\left| p \right|$}, & \text{if } \alpha \in (0,\pi) \text{ and + vertical intercept} \\
    -\textit{$\left| p \right|$}, & \text{if } \alpha \in (0,\pi) \text{ and - vertical intercept}\\
    0, & \text{if $q$ passes through the origin}  
\end{cases} \label{eq: p} 
\end{align}

\noindent where $\left| p \right|$ represents the shortest distance from the trajectory to the origin. An example of this transformation is illustrated in Fig. \ref{fig:1}.

We can also express the transformation of line $q$ in terms of its slope $m$ and vertical intercept 
$b$. This transformation, 
$T$, maps the equation of the line $q$ to a corresponding point 
$l$ with respect to parameters 
$\alpha$ and $p$, and can be formally written 
\begin{align*}
    T:& y  =mx+b  \mapsto  \left(\alpha = \frac{\pi}{2}+\arctan(m), p= \frac{b}{\sqrt{1+m^2}} \right) . 
\end{align*}
The inverse transformation $T^{-1}$ converts from $(\alpha,p)$ back to the line 
\begin{align*}
    T^{-1}:& (\alpha,p) \mapsto y= \tan \left(\alpha-\frac{\pi}{2}\right)x +p\sqrt{1+\tan^2  \left(\alpha-\frac{\pi}{2}\right)}.
\end{align*}
\noindent The transformation $T$ is designed to map each distinct line trajectory to a unique point, ensuring \emph{bijectiveness}, which is needed for sensor locations to be mapped uniquely between the inertial and representation spaces. It is straightforward to show that this transformation is bijective, which ensures that for each line in the inertial space, there is a unique point in the representation space. For each point in the representation space, there is a unique line in the inertial space.

\subsection{Stochastic linear target trajectory model: \\ log-Gaussian Cox line process (LGCLP)}

We define a line process passing through a convex compact set \( A \subset \mathbb{R}^2 \) in the inertial space by mapping the line process to a corresponding point process on \( \Theta \subset \mathcal{C} \).
To achieve this, from estimated historical linear trajectory data, we map the corresponding points in \( \Theta \) and then estimate the spatially-varying intensity function \( \lambda(l), \, l \in \Theta \), for the point process. For example, in Fig.~\ref{fig:2}, the red circle and box represent \( A \) and \( \Theta \), respectively, and the lines passing through \( A \) are mapped into the distinct points within the red box.

The time-period $T_c$ over which historical linear target trajectories are recorded is inherited by the intensity function, which is in units of density per unit time. Thus, the intensity of the Poisson line process mapped in $\Theta \subset \mathcal{C}$ is 
 \begin{align*}
     \Lambda(\Theta) = \frac{1}{T_c}\int_{\Theta} \lambda(l) dl
 \end{align*}
 To model the uncertainty that can arise when using historical data is used to estimate future target trajectories, we assume that the logarithm of $\lambda(l)$ is a Gaussian process. We note that the LGCP model is such that the intensity function is always non-negative. The intensity function is estimated from the historical data using the integrated nested Laplace approximation (INLA) method \cite{martino2019integrated,bachl2019inlabru, rue2009approximate}.
 
%%%%%%%%%%%%%%%%%%%%%%%%%%%%%%%%%%%%%%%%%%%%%%%%%%%%%%%%%%%%%%%%%%%%%%%%%%%%%%%%%%%%%%%%%%
\begin{figure}[t!] % figure 3
\centering
\includegraphics[scale=0.55]
{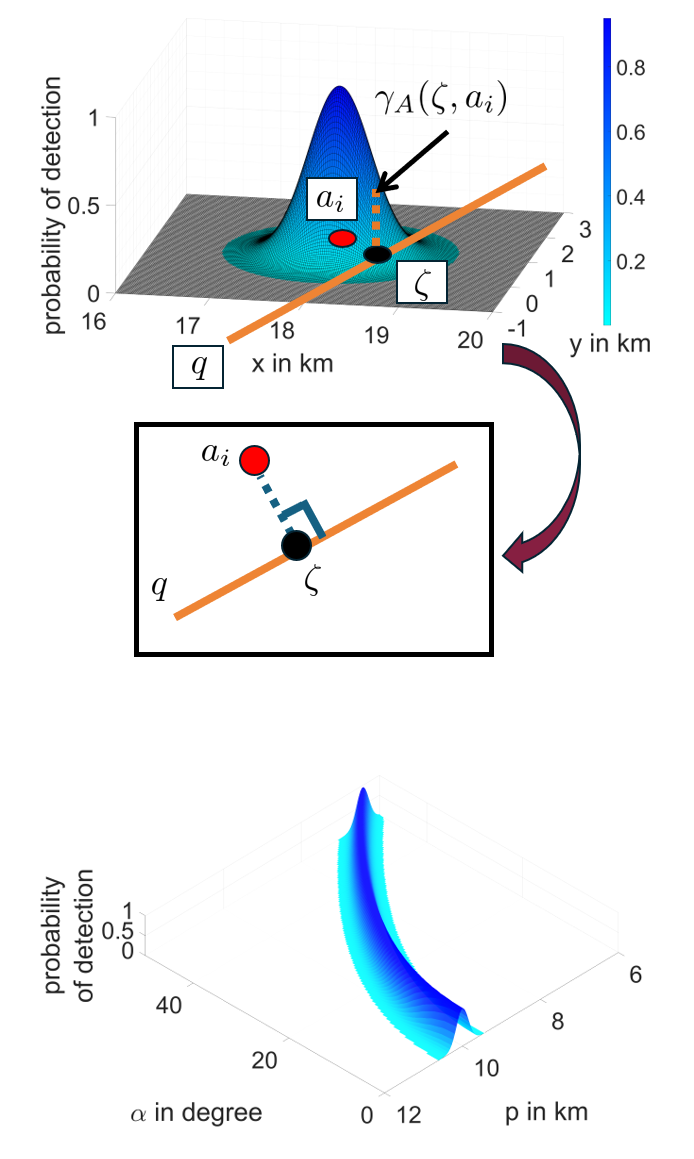}
\caption{(top) Visualized sensor performance function $\gamma_A$ (probability of detection of a linear trajectory $q$ with respect to  $\zeta$ and $a_i$). (bottom) The corresponding mapped sensor performance $\gamma_\mathcal{C}$ in the representation space.
}\label{fig:3}
\end{figure}
%%%%%%%%%%%%%%%%%%%%%%%%%%%%%%%%%%%%%%%%%%%%%%%%%%%%%%%%%%%%%%%%%%%%%%%%%%%%%%%%%%%%%%%%%%%%%%%%%%%%%%

%%%%%%%%%%%%%%%%%%%%%%%%%%%%%%%%%%%%%%%%%%%%%%%
\begin{figure*}[t!] % figure 4
\centering
\includegraphics[scale=0.43]
{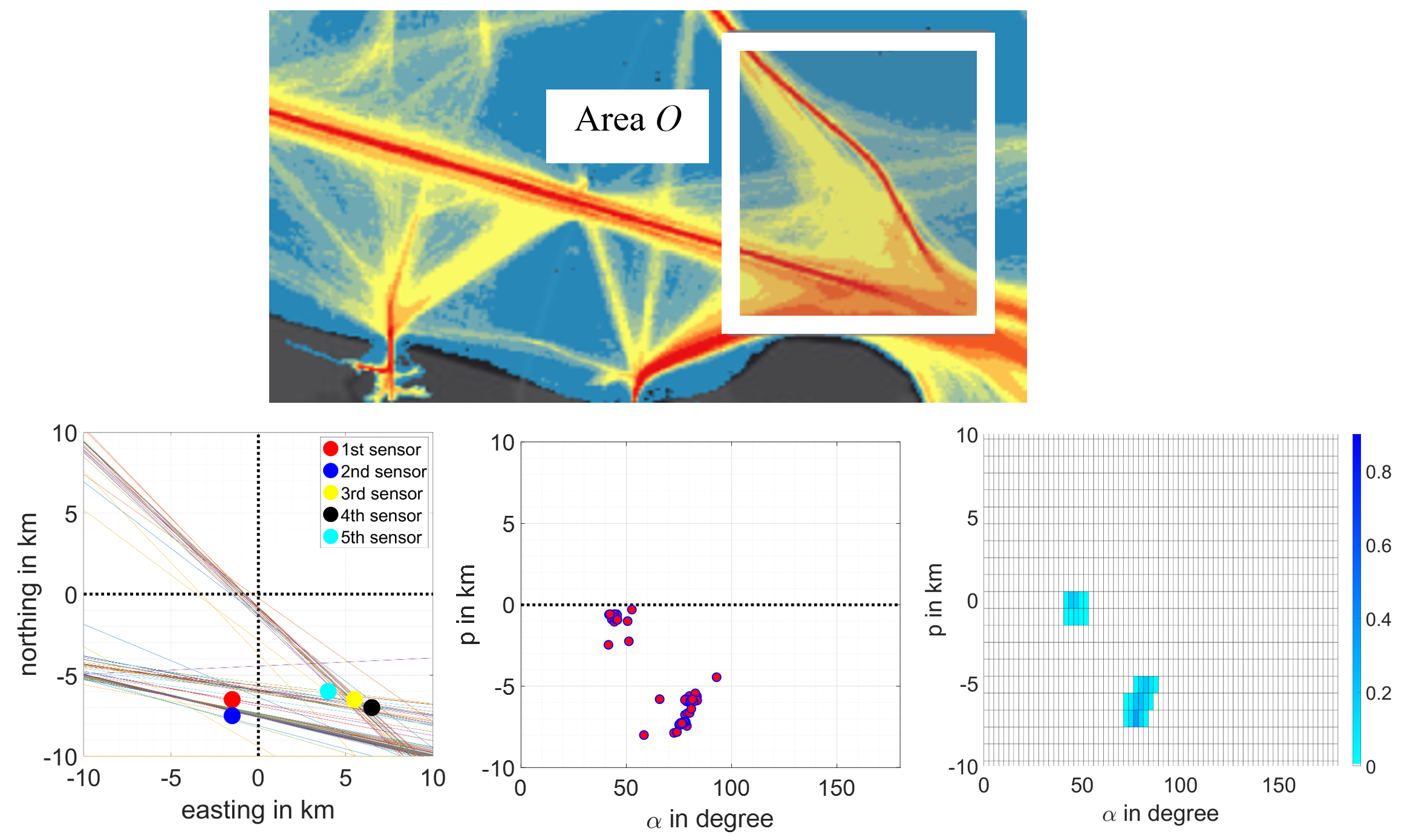}
\caption{Procedure of stochastic target trajectory estimation - (top) Heatmap of target traffic data \cite{marinecadastre.gov} around an area of interest $O$. (bottom-left) Estimated linear target trajectories within an area of interest with greedily selected 5 sensor locations. (bottom-center) Mapped unique points into the representation space using the linear trajectories from the (bottom-left). (bottom-right) Estimated mean intensity function of the mapper point pattern of the (center) in the representation space.
}\label{fig:4}
\end{figure*}
%%%%%%%%%%%%%%%%%%%%%%%%%%%%%%%%%%%%%%%%%%%%%%%%%%%%%%%%%%%%%%%%%%%%%%%%%%%%%%%%%%%%%%%%%%%%%%%%%%%%%%%%%%%%%%%%%%%%%%%%%%%%%%%%%

\subsection{Sensor model }\label{sec: sensor model}

For simplicity, we assume the sensor's detection probability is isotropic and decreases with distance. However, our method can be adapted for anisotropic cases.

We denote by $\zeta$ the location on a linear trajectory $q$ where the minimum distance to a sensor at location $a_i\in A$ is achieved in the inertial space (see the top figure in Fig. \ref{fig:3}). For each linear trajectory $q$ and each sensor at $a_i$, there is a unique $\zeta$. We define $\gamma_{A}(\zeta, a_{i}): A \times A  \rightarrow [0,1]$ as the probability of sensor $i$ detecting a target traveling on trajectory $q$. In other words, the probability of detection is determined by the minimum distance between the sensor location and the linear trajectory. The probability of failing to detect a target trajectory passing $q$ with a sensor at $a_i$ is expressed $ 1 - \gamma_{A}(\zeta, a_{i})$. Let $\mathbf{a} = \{a_{1}, a_{2}, \ldots, a_{M} \}$ denote the locations of a set of $M$ sensors in $A$. Then, the probability of failing to detect a target trajectory $q$ by the sensor network $\mathbf{a}$
\begin{align*}
     \pi_{A}(\zeta, \mathbf{a}):= \prod_{i=1}^{M} \big ( 1 - \gamma_{A}(\zeta, a_{i}) \big )
\end{align*}
Since we estimate the target trajectory intensities in $\mathcal{C}$, we need to map the probability of detection, $\gamma_{A}(\zeta, a_{i})$ into $\mathcal{C}$, yielding a probability of detection $\gamma_\mathcal{C}(l, a_i)$ for a unique point $l \in \mathcal{C}$. For a fixed value of $a_i$, we compute $\gamma_\mathcal{C} (l, a_i)$, which represents the probability that a sensor at $a_i$ detects the point $l$ in $\mathcal{C}$, corresponding to a unique line in the inertial space (see the bottom figure in Fig.\ref{fig:3}). Thus, the probability of failing to detect the target trajectory $l \in \mathcal{C}$ by the sensor network $\mathbf{a}$ is 
\begin{align*}
    \pi_{\mathcal{C}}(l, \mathbf{a}):= \prod_{i=1}^{M}  \big ( 1 - \gamma_{\mathcal{C}}(l, a_{i} )  \big )
\end{align*}

%%%%%%%%%%%%%%%%%%%%%%%%%%%%%%%%%%%%%%
\section{Suboptimal Sensor Placement: Maximizing Void probability Approximation}
\label{sec:sensorplacement}
%%%%%%%%%%%%%%%%%%%%%%%%%%%%%%%%%%%%%%

We place sensors at locations that maximize void probability, which is the probability of undetected targets being zero. Following the approach and analysis in \cite{kim2023toward}, which we briefly summarize in this section, sensors can be placed one at a time to greedily maximize an appropriate objective function. 

We denote the number of undetected target trajectories passing through \( A \) in the inertial space, given the sensor network \( \mathbf{a} \), by \( \bar{N}(A) \). The probability that no target trajectories are undetected, i.e., \( \bar{N}(A) = 0 \), given a target model \(\hat{\lambda} \) in the inertial space, is expressed as \( P(\bar{N}(A) = 0 \mid \hat{\lambda}) \).
Equivalently, we can express the same probability in the corresponding subset of the representation space $\mathcal{C}$. Let us denote $\Theta,\lambda(l)$ as the corresponding subset, intensity function in $\mathcal{C}$, respectively. By further marginalizing out the corresponding intensity function of the representation space yields the void probability that we seek to maximize,
\begin{align}
      P\left( \bar{N}\left( \Theta \right) = \right.  \left. 0 \right) 
      =\mathbb{E}_{\lambda} \left[e ^{-\int_{\Theta} \frac{1}{T_c} \lambda(l)   \pi_{\mathcal{C}}(l, \mathbf{a}) dl} \right] \label{eq:thinnedVoidprobability}
\end{align}
where the term $\int_{\Theta} \frac{1}{T_c} \lambda(l)  dl$ represents the number of targets within the bounded space $\Theta$ per time period $T_c$. By multiplying the integrated inside the integral by $\pi_{\mathcal{C}}(l, \mathbf{a})$, the probability of detection failure, we obtain the number of undetected targets $\int_{\Theta} \frac{1}{T_c} \lambda(l)   \pi_{\mathcal{C}}(l, \mathbf{a}) dl$. Intuitively, as we increase the number of sensors, the quantity $\int_{\Theta} \frac{1}{T_c} \lambda(l) \pi_{\mathcal{C}}(l, \mathbf{a}) dl$ decreases—a process we refer to as \emph{thinning}. Thus, optimal sensor placement in our model effectively means achieving the best thinning of the intensity curve to minimize the number of undetected targets. However, finding the optimal sensor locations of \eqref{eq:thinnedVoidprobability} is NP-hard \cite{krause2008near}, which motivates the use of a suboptimal solution.
We seek sensor locations $\hat{\mathbf{a}}  = \{\hat{a}_{1}, \ldots, \hat{a}_{M}\}$ that satisfy
\begin{align}
    \hat{\mathbf{a}} = \argmax_{\mathbf{a}} \nu(\mathbf{a})= e^{ -\int_{\Theta} \frac{1}{T_c}\mathbb{E}_{\lambda} \left[\lambda(l) \right]\pi_{\mathcal{C}}(l, \mathbf{a}) dl} \label{eq: greedy}
\end{align}
where \eqref{eq: greedy} is a lower bound for  \eqref{eq:thinnedVoidprobability} derived via Jensen's inequality. We show that greedily selected sensor locations for objective function in \eqref{eq: greedy} results in a value of the reward function that is at least $(1-1/e)$ of optimal (details appear in \cite{kim2023toward}, Appendix A).

As explained in Section IV of \cite{kim2023toward}, the difference in value between void probability in \eqref{eq:thinnedVoidprobability}
and void probability approximation in \eqref{eq: greedy} for a given set of sensor locations, referred to as Jensen's gap, is analyzed and shown to be small across a range of numerical examples. The Jensen's gap analysis outlined in \cite{kim2023toward} is applicable to this work. Due to its similarity, we omit the detailed derivation in this paper. In Section \ref{sec: numerical result}, we present the Jensen's gap for sensor locations determined not only by greedy selection but also by iterative nonlinear solvers, using the greedily selected sensor locations as the initial condition for \eqref{eq: greedy}. 

%%%%%%%%%%%%%%%%%%%%%%%%%%%%%%%%%%%%%%
\section{Numerical Result}\label{sec: numerical result}
%%%%%%%%%%%%%%%%%%%%%%%%%%%%%%%%%%%%%%

%%%%%%%%%%%%%%%%%%%%%%%%%%%%%%%%%%%%%%%%%%%%%%%%%%%%%%%%%%%%%%%%%%%%%%%%%%%%%%%%%%%%%%%%%%%%%%%%%%%%%%
\begin{figure}[t!] % figure 5
\centering
\includegraphics[scale=0.46]
{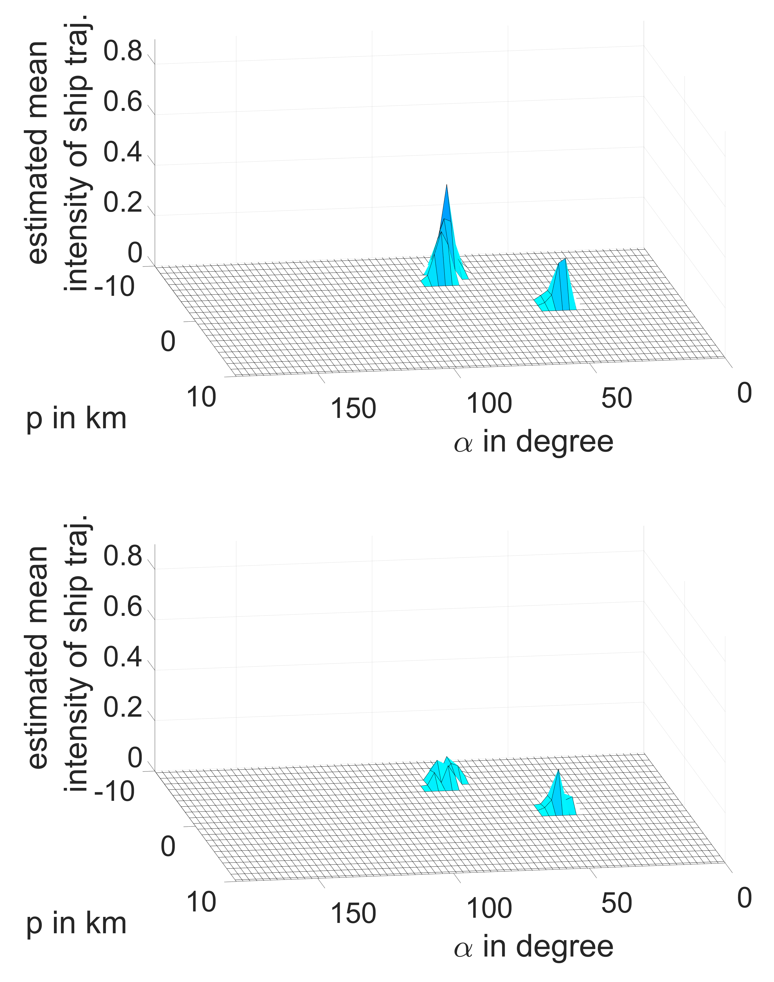}
\caption{(top) Estimated mean intensity function of LGCP modeled target trajectories. (bottom) Thinned the mean intensity function from (top) by placing a sensor performance.
}\label{fig:5}
\end{figure}
%%%%%%%%%%%%%%%%%%%%%%%%%%%%%%%%%%%%%%%%%%%%%%%%%%%%%%%%%%%%%%%%%%%%%%%%%%%%%%%%%%%%%%%%%%%%%%%%%%%%%%

In this section, we use historical ship traffic data to model stochastic linear target trajectories passing through a bounded area of interest. We generate a numerical example using ship location data obtained from the Automated Identification System (AIS) historical records near Hampton, Virginia, USA, sourced from \cite{marinecadastre.gov}. The data consists of ship identification, location, and time. Using these ship data, and by employing linear regression, we derive linear trajectories that approximate the ship paths represented in the AIS data.

The top figure in Fig.~\ref{fig:4} shows a heat map of the ship traffic data from January to May in 2022, where red indicates high-traffic regions, yellow represents areas of lower traffic, blue indicates no traffic, and gray denotes land. The white box in the heat map represents the area of interest, denoted as $O$. This area is defined by a latitude range from 36.91676 to 37.10289 and a longitude range from -76.10827 to -76.0904, both of which are simplified to a span of -10 km to 10 km in the northing and easting directions. The bottom-left figure in Fig.~\ref{fig:4} shows the estimated linear trajectories of the ship traffic within the area of \( O \) in March 2022.

Fig.~\ref{fig:4} illustrates the procedure for estimating the intensity of stochastic target trajectories in the representation space. This approach enables us to mathematically express the interaction between stochastic linear trajectories and sensor performance, allowing us to optimize sensor placement by thinning the estimated intensity function of the LGCP-modeled target trajectories in the representation space. To achieve this, we first map the estimated linear trajectories from the historical ship traffic data to corresponding unique points in the representation space, as shown in the transition from the bottom-left to the bottom-center figure in Fig.~\ref{fig:4}. Next, using the mapped points, we apply the integrated nested Laplace approximation (INLA) to estimate a log-Gaussian intensity function $\lambda(l)$, with intervals of 1 km for \( p \) and \( 2.5^{\circ} \) for \( \alpha \). The mean of the estimated intensity function is shown in the bottom-right figure of Fig.~\ref{fig:4}.

Subsequently, the process of searching for optimal sensor locations involves mapping the sensor performance (probability of detection) from the inertial space, as shown in Fig. \ref{fig:3} (top: cone shape), to the representation space, shown in Fig. \ref{fig:3} (bottom: tunnel shape). In the representation space, using this mapped sensor performance, we identify sensor locations that optimally thin the intensity function of the LGCP, maximizing the void probability. In our example, the sensor model in the inertial space is defined 
\begin{align}
    \gamma_{A}(\zeta,a) = \rho \exp(-((x-a_x)^2+(y-a_y)^2)/\sigma_l) \label{eq: sensor model numerical result}
\end{align} 
where $\zeta=\{x,y\}\in A$, $a = \{a_x,a_y\}\in A$, and $\rho$ and $\sigma_l$ denote the maximum probability of detection of a sensor and the length scale parameter, respectively. As described in Section \ref{sec: sensor model} using the shortest distance between a sensor location and a unique linear trajectory, when the sensor performance \eqref{eq: sensor model numerical result} is mapped into the representation space $\mathcal{C}$, the sensor model in the representation space is expressed
\begin{align*}
    \gamma_{\mathcal{C}}(l,a)= \rho \exp \left(- \left( p- \frac{ a_y + a_x/\tan \alpha  }{\sqrt{1+1/\tan^2\alpha}} \right)^2/\sigma_l\right)
\end{align*}
where $l=\{\alpha, p \}\in \mathcal{C}$. The term inside the square in the exponent represents the minimum distance between the sensor location $a$ and the line parameterized by ${p, \alpha}$. For our example, we use the parameter values $\rho = 0.95$ and $\sigma_l = 0.15$.

\begin{algorithm} [t!] 
  \caption{Sensor deployment for two-dimensional barrier coverage system: greedy selection}\label{algo: greedy}
  \begin{algorithmic}[1] 

\State \textbf{Input:} $ \mathbb{E}_{\lambda}[\lambda(l)]$, $ \mathbf{b}=\{b_1,...,b_W \}$ all possible discrete sensor locations where $b_j \in A$ and $j=\{1,...,W\}$

\State \textbf{Output:} $\hat{\mathbf{a}}=\{\hat{a}_1,...,\hat{a}_M\}$ sensor locations

\State \textbf{Initialize:} $\nu_{MAX}=0 $ \Comment{void prob. maximum},

\For{i=1:M} \Comment index for sensor $i$
    \For{j = 1:W} \Comment index for possible sensor location
        \State Map performance of a sensor placed at $b_j$ in $\mathcal{C}$
        \State Compute void prob. approx. $\nu(b_j)$ from \eqref{eq: greedy}
        \If{$\nu_{MAX} < \nu(b_j)$}  \Comment greedy selection
            \State $a_{MAX} = b_j$  
            \State $\nu_{MAX} = \nu(b_j)$
        \EndIf
    \EndFor
    \State $a_{MAX} \rightarrow  \hat{a}_i$\Comment Assigning $a_i$ 
    \State $\nu_{MAX}=0$ \Comment Resetting $\nu_{MAX}$

\EndFor

  \end{algorithmic}
\end{algorithm}
%%%%%%%%%%%%%%%%%%%%%%%%%%%%%%%%%%%%

% %%%%%%%%%%%%%%%%%%%%%%%%%%%%%%%%%%%
% %%%%%%%%%%%%%%%%%%%%%%%%%%%%%%%%%%%
\begin{table}[t!]
\caption{Void probability approximation \eqref{eq: greedy} with greedy selection and nonlinear iterative methods (Newton, quasi-Newton, and trust region methods) and the corresponding void probability.}
\begin{center}
\begin{tabular}{|c|c|c|c|}
\hline

\textbf{ \thead{$\#$ of \\ sensors}} & \textbf{ \thead{Void  prob. \\ approx.}}& \textbf{ \thead{Void  prob. }}& \textbf{\thead{Solver}}  \\
\hline

1 & 0.093 & 0.096 & \thead{greedy } \\
\hline
2 & 0.164 & 0.167 & \thead{greedy } \\
\hline
3 & 0.242 & 0.246 & \thead{greedy } \\
\hline
4 & 0.301 & 0.306 & \thead{greedy } \\
\hline
5 & 0.364 & 0.368 & \thead{greedy } \\
\hline
5 & 0.365 & 0.369 & \thead{greedy  + Newton } \\
\hline
5 & 0.372 & 0.374 & \thead{greedy  + quasi-Newton } \\
\hline
5 & 0.369 & 0.373 & \thead{greedy  +  trust region} \\
\hline

\end{tabular}
\label{tab1}
\end{center}
\end{table}

%%%%%%%%%%%%%%%%%%%%%%%%%%%%%%%%%%%
%%%%%%%%%%%%%%%%%%%%%%%%%%%%%%%%%%%

% % %%%%%%%%%%%%%%%%%%%%%%%%%%%%%%%%%%%
% % %%%%%%%%%%%%%%%%%%%%%%%%%%%%%%%%%%%
% \begin{table}[t!]
% \caption{Void probability approximation \eqref{eq: greedy} with greedy selection and nonlinear iterative methods (Newton, quasi-Newton, and trust region methods).}
% \begin{center}
% \begin{tabular}{|c|c|c|}
% \hline

% \textbf{ \thead{$\#$ of \\ sensors}} & \textbf{ \thead{Void  probability \\ approximation}}& \textbf{\thead{Solver}}  \\
% \hline

% 1 & 0.009 & \thead{greedy selection} \\
% \hline
% 2 & 0.046 & \thead{greedy selection} \\
% \hline
% 3 & 0.224 & \thead{greedy selection} \\
% \hline
% 4 & 0.369 & \thead{greedy selection} \\
% \hline
% 5 & 0.556 & \thead{greedy selection} \\
% \hline
% 5 & 0.561 & \thead{greedy selection + Newton } \\
% \hline
% 5 & 0.565 & \thead{greedy selection + quasi-Newton } \\
% \hline
% 5 & 0.569 & \thead{greedy selection +  trust region} \\
% \hline

% \end{tabular}
% \label{tab1}
% \end{center}
% \end{table}

% %%%%%%%%%%%%%%%%%%%%%%%%%%%%%%%%%%%
% %%%%%%%%%%%%%%%%%%%%%%%%%%%%%%%%%%%
To greedily search for a suboptimal set of sensors in the representation space that maximizes the void probability, we discretize the potential sensor locations within the area of interest in the inertial space. We select sensor locations for an optimization problem over a discrete domain by discretizing the inertial space. For this example, we create a grid ranging from -10 km to 10 km on both the easting and northing axes, with intervals of 0.5 km between points. 

As shown in Algorithm 1, to select sensor locations with the greedy approach, we determine a sensor location that maximally thins the estimated mean intensity function, \(\mathbb{E}_{\lambda}[\lambda(l)]\), among all possible options, following \eqref{eq: greedy}. The three-dimensional estimated mean intensity function \(\mathbb{E}_{\lambda}[\lambda(l)]\) is shown in the top of Fig.~\ref{fig:5}, which corresponds to the bottom-right image in Fig.~\ref{fig:4}. In Fig.~\ref{fig:4} (bottom-left), the small red circle marks the location that initially thins the intensity function \(\mathbb{E}_{\lambda}[\lambda(l)]\) most. The bottom image in Fig.~\ref{fig:5} displays the thinned estimated mean intensity after applying the first greedily chosen sensor location to \(\mathbb{E}_{\lambda}[\lambda(l)]\). By repeating this process, the resulting five greedily selected sensor locations are shown as small, differently colored circles in Fig.~\ref{fig:4} (bottom-left), positioned within the inertial space alongside the historical ship trajectories. The void probability achieved by greedily placing each of the first 5 sensors is shown in the first 5 rows of Table \ref{tab1}.

% Furthermore, the results, detailed in the top five rows in the second column from the left of Table \ref{tab1}, show that the void probability approximation increases by an average of 0.073. 

\subsection*{Jensen's gap }

We greedily search for the sensor locations that maximize the void probability approximation from \eqref{eq: greedy} shown in the second column from the left in Table \ref{tab1}. With this set of sensor locations, we can estimate the actual void probability using Monte Carlo sampling 

\begin{align*}
    \mathbb{E}_{\lambda} \left[e ^{-\int_{\Theta} \frac{1}{T_c} \lambda(l)   \pi_{\mathcal{C}}(l, \mathbf{a}) dl} \right] \approx \frac{1}{Z} \sum_{k=1}^Z e ^{-\int_{\Theta} \frac{1}{T_c} \Tilde{\lambda}_k(l)   \pi_{\mathcal{C}}(l, \mathbf{a}) dl} 
\end{align*}
where $Z$ is the number of intensity function samples, and $\Tilde{\lambda}_k(l)$ is $k^{th}$ sampled intensity function. For this analysis, we use $Z=10,000$ samples. In our example, the Jensen gap appears small, with an average gap value of 0.0038.

\subsection*{Optimized sensor locations using nonlinear iterative techniques}

Specifically, we evaluate Newton, quasi-Newton, and trust region methods—commonly applied nonlinear iterative optimization techniques \cite{NoceWrig06}. Beginning with an initial set of five sensor locations selected via a greedy algorithm, the nonlinear optimization algorithms improve the approximate void probability by approximately 0.90$\%$, 2.34$\%$, and 1.62$\%$, respectively. The nonlinear optimization algorithms are applied following Algorithm 2, where step 10 is specifically tailored for each optimization method, adjusting step size and direction iteratively using the gradient and Hessian of the objective function as required by the nonlinear optimization algorithm. This approach allows for more precise location adjustments, as each method refines the step size and direction at each iteration based on the local curvature information provided by the Hessian.

In our problem, the objective function is non-convex with respect to sensor locations, making it challenging to identify the optimal positions. To address the non-convexity of our objective function in \eqref{eq:thinnedVoidprobability}, we approximate the Hessian as positive definite (step 8 in Algorithm 2), ensuring that the optimization methods can increase the void probability more effectively. By leveraging curvature information through both gradient and modified Hessian, Newton, quasi-Newton, and trust region methods provide a pathway toward global convergence, allowing us to achieve a more robust optimization of sensor placement \cite{NoceWrig06}.

\begin{algorithm} [t!] 
  \caption{Sensor deployment for two-dimensional barrier coverage system: greedy selection $+$ nonlinear iterative optimization methods}\label{algo: steepestGradientAscent}
  \begin{algorithmic}[1] 

\State \textbf{Input:} $\epsilon$ stopping criteria (small), $\mathbb{E}_{\lambda}[\lambda(l)]$, $ \gamma_{\mathcal{C}}(l,a)$

\State \textbf{Output:} 
$\Tilde{\mathbf{a}}=\{\Tilde{a}_1,...,\Tilde{a}_M\}$ refined sensor locations

\State \textbf{Initialize:} $\mathbf{a} = \hat{\mathbf{a}}$ greedily selected sensor locations from Algorithm 1

\While {$\epsilon < $ norm of gradient of $\nu(\mathbf{a})$}
    \State Compute gradient of $\nu({\mathbf{a}})$
    \State Compute approximated or full Hessian of $\nu({\mathbf{a}})$

    \If{ Hessian is indefinite}
        \State
        \parbox[t]{\dimexpr\linewidth-\algorithmicindent-\algorithmicindent}{%
           Apply modified symmetric indefinite factorization from \cite{NoceWrig06} to 
           the computed indefinite Hessian.\\%
    }
    \EndIf
    
    \State
    \parbox[t]{\dimexpr\linewidth-\algorithmicindent-\algorithmicindent}{%
     Compute step size/direction of sensor locations using the computed gradient/Hessian. \\%
     }
    % \State Compute step size/direction of sensor locations
    
    \State
    \parbox[t]{\dimexpr\linewidth-\algorithmicindent-\algorithmicindent}{%
    Update sensor locations  $\Tilde{\mathbf{a}} $ using computed step size/direction.\\ %
    }

\EndWhile

  \end{algorithmic}
\end{algorithm}

%%%%%%%%%%%%%%%%%%%%%%%%%%%%%%%%%%%%%%
\section{Conclusion}\label{sec: conclusion}
%%%%%%%%%%%%%%%%%%%%%%%%%%%%%%%%%%%%%%

In conclusion, we investigate suboptimal sensor placement for a barrier coverage problem aimed at detecting Poisson-distributed linear trajectories in a two-dimensional domain. Our approach represents these trajectories as unique points within a transformed representation space, facilitating effective estimation of the intensity of stochastic linear target trajectories. By mapping sensor performance onto this space, we identify suboptimal sensor locations that maximize the probability of detecting all targets, achieving a high probability of perfect detection. Numerical experiments using historical ship data validate the effectiveness of our approach in accurately estimating and detecting uncertain linear trajectories of targets.\\

\bibliographystyle{IEEEtran}
\bibliography{main.bib}

\end{document}